\documentclass{article}
\usepackage[preprint]{colm2026_conference}

\usepackage{amsmath,amssymb,amsthm}
\usepackage{graphicx}
\usepackage{booktabs}
\usepackage{xcolor}
\usepackage{subcaption}
\usepackage{float}
\usepackage{multirow}
\usepackage{microtype}
\usepackage{url}
\usepackage{longtable}
\usepackage{lineno}

\newcommand{\R}{\mathbb{R}}

\title{Finding Belief Geometries with Sparse Autoencoders}

\author{
  Matthew Levinson\thanks{This work was supported by a scholarship and
  computational resources from Simplex AI Safety. The author thanks Paul
  Riechers, Adam Shai, and Xavier Poncini for their support.} \\
  Independent Researcher \\
  \texttt{good.epic@gmail.com}
}

\date{}

\begin{document}

\ifcolmsubmission
\linenumbers
\fi

\maketitle

\begin{abstract}

  Understanding the geometric structure of internal representations is a central
  goal of mechanistic interpretability. Prior work has shown that transformers
  trained on sequences generated by hidden Markov models encode probabilistic
  belief states as simplex-shaped geometries in their residual stream,
  with vertices corresponding to latent generative states.
  Whether large language models trained on naturalistic text develop analogous
  geometric representations remains an open question.

  We introduce a pipeline for discovering candidate simplex-structured subspaces
  in transformer representations, combining sparse autoencoders (SAEs),
  $k$-subspace clustering of SAE features, and simplex fitting using AANet.
  We validate the pipeline on a transformer trained on a multipartite hidden
  Markov model with known belief-state geometry. Applied to Gemma-2-9B, we
  identify 13 priority clusters exhibiting candidate simplex geometry ($K \geq 3$).

  A key challenge is distinguishing genuine belief-state encoding from
  \emph{tiling artifacts}: latents can span a simplex-shaped
  subspace without the mixture coordinates carrying predictive signal beyond
  any individual feature. We therefore adopt \emph{barycentric prediction}
  as our primary discriminating test. Among the 13 priority
  clusters, 3 exhibit a highly significant advantage on near-vertex samples
  (Wilcoxon $p < 10^{-14}$) and 4 on simplex-interior samples. Together
  5 distinct real clusters pass at least one split, while no null cluster
  passes either. One cluster, 768\_596, additionally achieves the highest
  causal steering score in the dataset. This is the only case where passive prediction
  and active intervention converge. We present these findings as preliminary
  evidence that genuine belief-like geometry exists in Gemma-2-9B's
  representation space, and identify the structured evaluation that would be
  required to confirm this interpretation.

\end{abstract}

\section{Introduction}
\label{sec:intro}

A growing body of work shows that transformer language models organize their
activations into structured geometric objects that encode variables relevant to
prediction \citep{elhage2022toy, cunningham2023sparse, bricken2023monosemanticity,
gurnee2023language}. One natural form such structure could take is a simplex. If
the model maintains a probability distribution over discrete latent states, the
barycentric coordinates of a simplex provide a natural internal representation of
that distribution. We refer to representations with this structure as \emph{belief
geometries}.

Prior work demonstrates this concretely in controlled settings. Transformers
trained on sequences from hidden Markov models develop simplex-shaped geometries
in their residual stream, with vertices corresponding to latent states and
barycentric coordinates encoding the model's belief distribution
\citep{shai2024transformers, piotrowski2025constrained}. Geometric structure
has also been identified in algorithmic tasks: addition encoded via trigonometric
representations \citep{kantamneni2025trigonometry}, and character counts on
curved manifolds \citep{gurnee2025manifolds}.
Whether analogous structures arise for the abstract latent variables of natural
language such as discourse mode, referential context, or syntactic role remains an
open question.

We introduce a pipeline for discovering candidate belief geometries in large
pretrained language models, combining SAE representations with $k$-subspace
clustering and AANet simplex fitting, a neural implementation of archetypal
analysis that recovers barycentric coordinates over extreme-point vertices
(Section~\ref{sec:real_aanet}). A central challenge is distinguishing
genuine belief-state encoding from \emph{tiling artifacts}: a group of
co-activating latents may span a simplex-shaped subspace simply by covering
different regions of a shared contextual variable, without encoding a genuine
mixture. Such tiling would produce apparent simplex geometry while the best
individual latent already captures the full predictive signal. We therefore
adopt the \emph{barycentric predictive advantage} as our primary test: genuine
mixture encoding implies that the full coordinate vector predicts next-token
behavior better than any single feature.

We first validate the pipeline on a controlled toy model where ground-truth
belief geometry is known. We then apply it to Gemma-2-9B, identifying 13
priority clusters as candidates for detailed evaluation. The barycentric
predictive advantage test is our primary discriminator: 3 of 13 real clusters
pass on near-vertex samples ($p < 10^{-14}$), 4 pass on simplex-interior
samples, and no null cluster passes on either split. Cluster 768\_596 is the
one case where passive prediction and active causal intervention jointly
support a belief-geometry interpretation. We present highlight results for
two clusters with the strongest causal and predictive signals, respectively,
and discuss what structured evaluation would be required to move from these
preliminary findings to a confirmed account of belief-state encoding in
naturalistic text.

\section{Related Work}
\label{sec:background}

Sparse autoencoders applied to transformer residual streams recover large
collections of sparse, interpretable feature directions
\citep{cunningham2023sparse, bricken2023monosemanticity}, providing a natural
basis for studying higher-order geometric structure. Beyond individual features,
structured geometric objects have been identified in transformer representations:
curved manifolds encoding numeric variables in algorithmic settings
\citep{gurnee2025manifolds, kantamneni2025trigonometry}, and linear subspaces
encoding more abstracted numeric concepts of physical and calendar distance
\citep{gurnee2023language}. Simplex representations
are of particular interest because they encode probability distributions
over discrete latent states through barycentric coordinates. Transformers have also
been shown to perform implicit Bayesian inference
\citep{akyurek2023icl, vonoswald2023transformers, xie2022meta-learning}, which
suggests belief geometries as a potential encoding strategy. Simplices are related
geometrically and conceptually to archetypal analysis \citep{cutler1994archetypal},
a statistical technique that looks for extreme points in a dataset from which other
examples can be represented as convex combinations. This makes it a good technique
for finding potential simplex vertices.

Most directly relevant is work showing that transformers trained on
HMM-generated sequences represent belief distributions geometrically as
simplices in the residual stream \citep{shai2024transformers,
piotrowski2025constrained}, where barycentric coordinates encode the model's
belief distribution over latent states. Our work extends this line by
introducing a method for discovering analogous structures in large pretrained
models without access to ground-truth latent states, and by introducing
the barycentric predictive advantage test to distinguish genuine mixture
encoding from tiling artifacts. A key difference from algorithmic settings is
that naturalistic discourse variables lack ground-truth labels, making
confirmation of the belief-state interpretation contingent on structured
evaluation infrastructure that does not yet exist, a gap we discuss in
Section~\ref{sec:future_validation}.

\section{Toy Model: Proof of Concept}
\label{sec:toy}

We validate the pipeline on a controlled multipartite hidden Markov model
with known ground-truth belief geometry (Appendix~\ref{app:toy}).
The model combines five independent generative components with a joint
vocabulary of 432 tokens, creating an intentionally challenging regime
where each token encodes contributions from all five sources simultaneously.
Applied to a small transformer trained on this model, the pipeline recovers
all five components with no supervision (mean $R^2 = 0.61$ on held-out data;
Table~\ref{tab:toy_r2} in Appendix~\ref{app:toy}), correctly identifying
the underlying simplex dimension for each component via the elbow criterion.

\section{Real Model Pipeline: Gemma-2-9B}
\label{sec:real}

In contrast to prior work that analyzes models trained on synthetic tasks or
real tasks with known latent variables, our approach searches for
simplex-structured representations in a large pretrained language model without
access to ground-truth latent states. We apply our belief-geometry discovery
pipeline to Gemma-2-9B, using a publicly released GemmaScope JumpReLU SAE for the
residual stream at layer 20 ($d_{\mathrm{SAE}} = 16{,}384$ latents;
\citealt{saelens2024}). The pipeline consists of four stages: 1) SAE
decoder directions are clustered to identify groups of features that may
jointly encode a single contextual variable; 2) we fit AANet to each
candidate group using FineWeb sequences \citep{penedo2024fineweb} to test for
simplex structure and recover barycentric coordinates; 3) we prioritize
clusters whose fitted geometry is most consistent with a $k$-simplex
($k \geq 3$); 4) we evaluate priority clusters using predictive, causal,
and semantic validation tests.

\subsection{$k$-Subspace Clustering}
\label{sec:real_cluster}

We cluster all column-normalized decoder directions in $W_d$ using a method
that combines $k$-subspace clustering \citep{wang2009ksub} with
column-pivoted QR initialization \citep{damle2018copr}. $k$-Subspace
clustering partitions vectors into groups that each span a low-dimensional
subspace, generalizing $k$-means from cluster centroids to cluster subspaces.
We seed the algorithm using column-pivoted QR decomposition of the decoder
matrix, which greedily selects a maximally spread set of initial subspace
directions and substantially improves convergence over random initialization.
We cluster decoder directions rather than token activations directly because
decoder geometry captures which latents write into similar residual-stream
subspaces, providing a model-based notion of features that may jointly encode
a common contextual variable.

We run the pipeline at two resolutions, $K \in \{512, 768\}$, and retain
candidate clusters with estimated rank $k \geq 3$.

\begin{figure}[t]
  \centering
  \includegraphics[width=\textwidth]{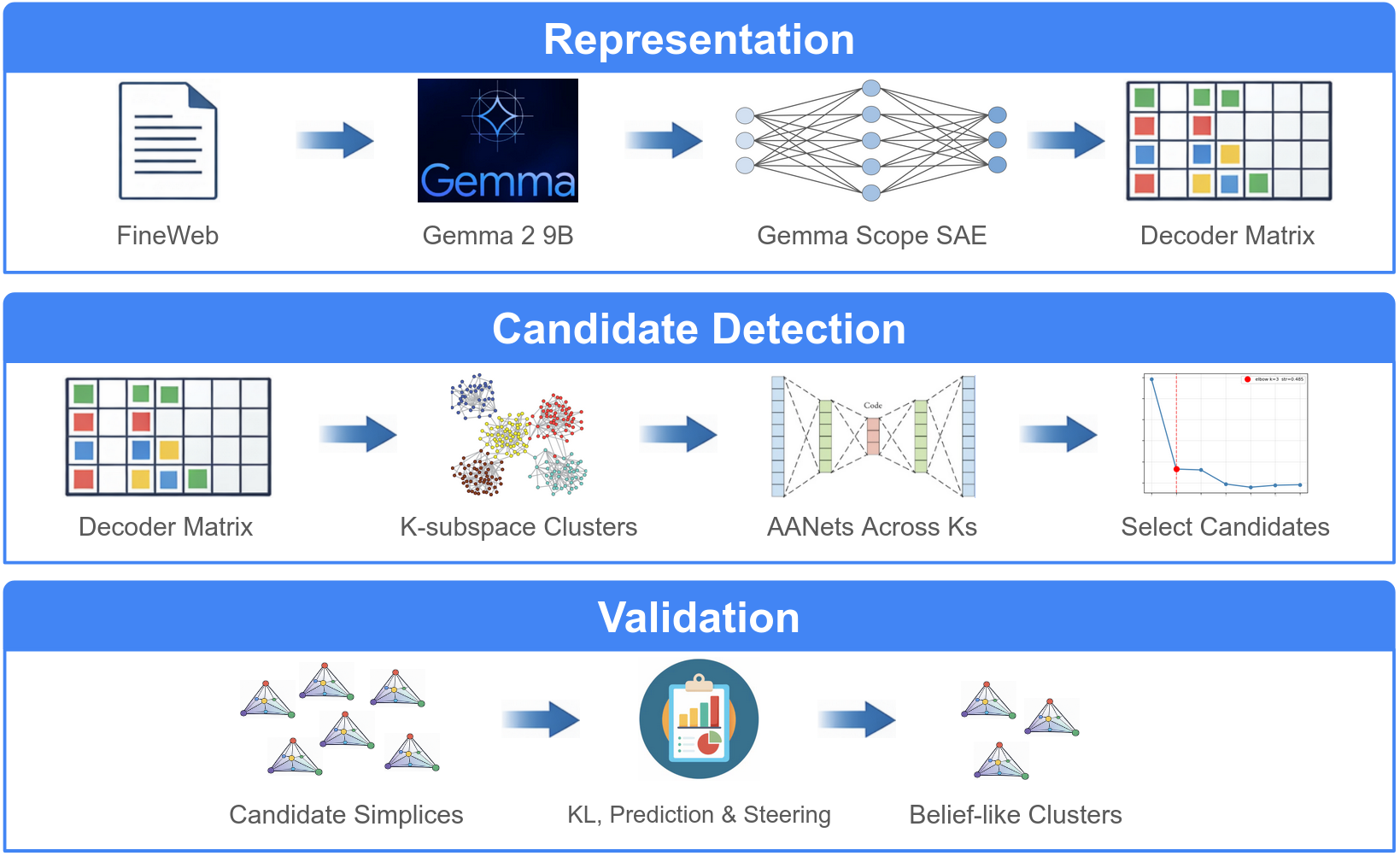}
  \caption{%
    \textbf{Overview of the belief geometry discovery pipeline.}
    Text sequences are processed through Gemma-2-9B, and residual-stream
    activations at layer 20 are encoded by a GemmaScope JumpReLU SAE. Decoder directions of
    SAE latents are clustered into candidate latent groups using
    $k$-subspace clustering. Each candidate group is then fit with AANet to test
    for simplex structure and recover barycentric coordinates. The resulting
    candidate belief geometries are evaluated using barycentric predictive
    advantage and causal steering as the primary validation analyses.
  }
  \label{fig:pipeline}
\end{figure}

\subsection{AANet Simplex Fitting}
\label{sec:real_aanet}

AANet \citep{vanDijk2019AAnet, vanDijk2024AAnet} is an autoencoder-based
implementation of archetypal analysis. It learns to embed data into a simplex
of $K$ vertices at the bottleneck layer, from which we extract barycentric
coordinates expressed as convex combinations of $K$ archetypal extreme points.
Unlike classical archetypal analysis, AANet learns a nonlinear encoder and
decoder jointly with the positions of the archetypes, making it robust to the
curved geometry expected in large model activations.

For each candidate cluster, we form the cluster's partial contribution to the
residual stream at each token position $t$ where at least one cluster latent
is active:
\begin{equation}
  \mathbf{x}_t = \tilde{\mathbf{f}}_t \,\mathbf{W}_d,
\end{equation}
where $\tilde{\mathbf{f}}_t$ is the SAE latent activation vector with all
non-cluster entries set to zero. We fit AANets for $K \in \{2, 3, 4, 5, 6, 7\}$
and identify clusters with an elbow in the reconstruction loss curve. We do not
search for cases with $K=2$ vertices, as a 1-simplex is not distinct from
a linear feature.

\subsection{Priority Cluster Selection}
\label{sec:real_selection}

\begin{table}[t]
  \centering
  \caption{%
    \textbf{The 13 priority clusters.}
    Clusters are identified by the number of clusters for the $k$-subspace
    clustering (512 or 768) and their cluster index.
    $m$ denotes the number of SAE latents in the cluster. Vertex consistency
    is labeled as consistent, mixed, or inconsistent.
    Confidence is derived from 20-iteration autointerp synthesis:
    HIGH $=\geq$2 consistent vertices;
    MEDIUM $=$ no inconsistent vertices;
    LOW $=\geq$1 inconsistent vertex.
  }
  \label{tab:priority_clusters}
  \begin{tabular}{lcl p{6.5cm}}
    \toprule
    Cluster & $m$ & Semantic Confidence & Vertex interpretation (summary) \\
    \midrule
    512\_17   & 15  & HIGH   & Text register (instructional / promotional / metadata) \\
    \midrule
    512\_229  & 22  & MEDIUM & Syntactic complexity (simple / formal-specific / complex-embedded) \\
    512\_261  & 18  & MEDIUM & Grammatical function (content words / function words / technical register) \\
    512\_504  & 22  & MEDIUM & Syntactic role (function words / content nouns / predicates) \\
    768\_140  & 26  & MEDIUM & Referential specificity (generic / specific / anaphoric) \\
    768\_210  & 4   & MEDIUM & Syntactic complexity (complex NP / simple / intermediate) \\
    768\_306  & 5   & MEDIUM & Register formality (conversational / formal-technical / structured) \\
    768\_596  & 6   & MEDIUM & Grammatical person / discourse role (3rd / 1st / 2nd person) \\
    \midrule
    512\_22   & 6   & LOW    & Grammatical role (nominal / function words / verbal) \\
    512\_67   & 25  & LOW    & Syntactic function (nominal-boundary / determiner-initiation / connective) \\
    512\_181  & 26  & LOW    & Syntactic role (content words / function words / specialized vocab) \\
    512\_471  & 10  & LOW    & Grammatical person / complexity (simple / third-formal / instructional) \\
    768\_581  & 11  & LOW    & Syntactic role (content words / function words / verbal predicates) \\
    \bottomrule
  \end{tabular}
\end{table}

After clustering and AANet fitting across both $N=512$ and $N=768$ runs, we
select 13 \emph{priority clusters} for detailed validation
(Table~\ref{tab:priority_clusters}). Eleven have an AANet elbow at $K=3$; two
have an elbow at $K=4$. Cluster sizes range from 4 to 26 latents.

For comparison, we construct \emph{null clusters}. We partitioned the latents
randomly into sets with the identical distribution of $m$ as the true clusters.
We retained the three that passed the same geometric screening criteria applied
to real candidates, then ran the same validation pipeline. The goal was to
test whether the pipeline over-selects spurious geometry. Note that even if we
loosened the criteria, only three null clusters passed the first screen. The 13-to-3
asymmetry at the screening stage is itself consistent with real clusters exhibiting
genuine low-dimensional structure.

\section{Validation}
\label{sec:validation}

We evaluate the 13 priority clusters and 3 null clusters using two main validation analyses.
A central challenge is the \emph{tiling artifact}.
Groups of latents may span a subspace that can be projected onto a simplex because they
cover different regions of a subspace leveraged by the model within a given context, without
encoding a genuine mixture. In that case, the best individual latent already captures the
full predictive signal, and the mixture coordinates add no value.

The barycentric predictive advantage is the primary discriminating test. It directly
addresses the tiling artifact, and its result on the null
clusters is the clearest single finding in the paper. Causal steering is the other main test.
When predictable results from active intervention can
be demonstrated, it is a stronger claim. In our case, we are searching for abstract latent
mixtures with an analytical setup that yields wide confidence intervals. So our steering
results are suggestive but not definitive. KL divergence and semantic coherence are
supporting analyses.

\subsection{Barycentric Predictive Advantage}
\label{sec:bary}

\paragraph{Setup.} The primary test asks whether the full barycentric coordinate
vector $\mathrm{bary}_t \in \R^{K}$ predicts next-token behavior better than any
single latent in the cluster. For each of the top-50 highest-variance tokens
across near-vertex samples, we fit a linear model predicting the token's
log-probability from the barycentric coordinates (5-fold cross-validation,
capped at 200 near-vertex samples per vertex). We separately fit the same
prediction using each individual latent activation. Using a paired Wilcoxon
signed-rank test, we compare the 50-vector of $R^2_{\mathrm{bary}}$ to the
50-vector of $R^2_{\mathrm{best\,latent}}$, where the latent $R^2$ vector uses
the highest $R^2$ across all individual latent models for each token.

\paragraph{Hypothesis.} Genuine mixture encoding implies that the full
barycentric representation should carry predictive signal not present in any
single latent, directly ruling out the tiling artifact.

\begin{table}[t]
  \centering
  \caption{%
    \textbf{Barycentric vs.\ best-latent predictive advantage, both evaluation splits.}
    \emph{Frac.\ bary wins}: fraction of top-50 highest-variance tokens for which
    the barycentric coordinate achieves higher $R^2$ than the best individual latent.
    Wilcoxon $p$-values test the one-sided hypothesis that barycentric $R^2$ is
    stochastically greater. NV = near-vertex samples; SI = simplex-interior samples.
    No null cluster passes on NV. One null cluster passes SI; see text.
  }
  \label{tab:bary}
  \small
  \begin{tabular}{lc cc cc}
    \toprule
    & & \multicolumn{2}{c}{Near-vertex (NV)} & \multicolumn{2}{c}{Simplex-interior (SI)} \\
    \cmidrule(lr){3-4} \cmidrule(lr){5-6}
    Cluster & $n$ & Frac.\ wins & $p$ & Frac.\ wins & $p$ \\
    \midrule
    512\_17   & 15 & 0.00 & 1.00          & 0.04 & 1.00 \\
    512\_22   &  6 & 0.00 & 1.00          & 0.04 & 1.00 \\
    512\_67   & 25 & 0.00 & 1.00          & 0.00 & 1.00 \\
    \textbf{512\_181}  & 26 & \textbf{1.00} & $<10^{-15}$ & \textbf{1.00} & $<10^{-15}$ \\
    \textbf{512\_229}  & 22 & \textbf{1.00} & $<10^{-15}$ & \textbf{0.58} & $0.003$ \\
    512\_261  & 18 & 0.00 & 1.00          & 0.16 & 1.00 \\
    \textbf{512\_471}  & 10 & 0.00 & 1.00 & \textbf{0.84} & $5{\times}10^{-9}$ \\
    \textbf{512\_504}  & 22 & 0.00 & 1.00 & \textbf{1.00} & $<10^{-15}$ \\
    768\_140  & 26 & 0.20 & 1.00          & 0.04 & 1.00 \\
    768\_210  &  4 & 0.00 & 1.00          & 0.00 & 1.00 \\
    768\_306  &  5 & 0.16 & 1.00          & 0.00 & 1.00 \\
    768\_581  & 11 & 0.00 & 1.00          & 0.00 & 1.00 \\
    \textbf{768\_596}  &  6 & \textbf{0.98} & $<10^{-14}$ & 0.16 & 1.00 \\
    \midrule
    512\_138 (null) & 51 & 0.00 & 1.00   & 0.00 & 1.00 \\
    512\_345 (null) & 47 & 0.06 & 1.00   & 0.00 & 1.00 \\
    768\_310 (null) & 17 & 0.22 & 0.977  & 0.00 & 1.00 \\
    \bottomrule
  \end{tabular}
\end{table}

\paragraph{Results.} Table~\ref{tab:bary} shows results for both evaluation
splits. No null cluster exhibits a significant barycentric advantage on either.
Among the 13 real clusters, 5 pass on near-vertex samples or interior samples,
with two passing on both. This directly tests the defining prediction of the
simplex hypothesis and rules out the tiling artifact for these clusters.

The simplex-interior split is the more informative test: near-vertex samples lie
close to a single extreme, so a specialized latent can already do well; interior
samples require the mixture coordinate to meaningfully shift token predictions,
something no single latent can accomplish for a true mixture. Clusters 512\_181
and 512\_229 pass both splits ($p \approx 0$), providing the strongest evidence.
Clusters 512\_471 and 512\_504 pass interior only; at the vertex extremes their
near-vertex $R^2$ values are nearly identical for barycentric and best-latent
predictions, consistent with a dominant latent at each extreme. Cluster 768\_596
passes near-vertex only (interior: frac.\ wins $= 0.16$, mean $R^2$: $-0.003$
vs.\ $0.093$), indicating its advantage is specific to the vertex extremes.
Together, 5 real clusters pass at least one split versus 0 null clusters on
either, a signal we take as evidence that the pipeline is finding real structure.

\paragraph{Distributional sanity check.} As a complementary check, we compute
the ratio of cross-vertex to same-vertex symmetric KL divergence between
next-token distributions at near-vertex positions. If vertex assignments are
functionally meaningful, cross-vertex pairs should predict more divergent
continuations. All 13 real clusters achieve ratios above 1 (1.017--1.987),
confirming that vertex assignments correspond to genuinely different
distributional regimes. However, null clusters also achieve ratios above 1
(1.108--1.329), so this check is not discriminating on its own; we treat it as
a sanity check that the near-vertex samples are at distributional extremes, not
as evidence for belief-geometry encoding. Full per-cluster values are in
Appendix~\ref{app:all_results}.

\subsection{Causal Steering}
\label{sec:steering}

\paragraph{Semantic prerequisite.} Interpreting a steering score requires first
establishing that the vertex labels are meaningful. For each cluster, we run
20 independent labeling rounds with Claude Sonnet 4.5 and measure \emph{document
accuracy} via Qwen-2.5-72B-AWQ classifying held-out near-vertex continuations
against the synthesized vertex labels (details in Appendix~\ref{app:steering_types}).
We require a mean score $\geq 0.15$ on at least 2 of $K$ vertices before a
cluster \emph{qualifies} for steering evaluation. Five real clusters do not
qualify; all three null clusters contain at least one vertex below this threshold,
though semantic coherence is not otherwise discriminating. Consolidated
interpretations are in Table~\ref{tab:priority_clusters} and
Appendix~\ref{app:all_results}.

\paragraph{Setup.} For qualifying clusters, we test whether the simplex
geometry is causally relevant by steering the model's residual stream toward
target vertex coordinates. For each source vertex $v$ and target $v' \neq v$,
we compute an intervention direction from the AANet vertex decoder, add it at
scale $s \in \{1, 5, 20\}$, and generate a continuation. We evaluate each
continuation using Qwen-2.5-72B (AWQ quantized, via vLLM), which classifies
whether the continuation matches the target vertex's semantic description. The
steering score is the mean classification accuracy across source-target pairs,
intervention types (Appendix~\ref{app:steering_types}), and scales.

\begin{table}[t]
  \centering
  \caption{%
    \textbf{Causal steering scores.}
    Null cluster scores overlap with those of most qualifying real clusters;
    768\_596 is the only cluster where a significant barycentric advantage and
    a positive steering score converge.
  }
  \label{tab:steering}
  \begin{tabular}{lccc}
    \toprule
    Cluster & Steering score & Best type & Best scale \\
    \midrule
    \textbf{768\_596} & \textbf{0.419} & type2 & 20 \\
    512\_22   & 0.356 & type1 & 20 \\
    768\_210  & 0.336 & type2 & 1 \\
    768\_140  & 0.289 & type1 & 20 \\
    512\_67   & 0.250 & type1 & 20 \\
    512\_261  & 0.242 & type1 & 1 \\
    768\_581  & 0.209 & type3 & 5 \\
    512\_17   & 0.118 & type3 & 5 \\
    \midrule
    512\_138 (null) & 0.358 & type1 & 1 \\
    768\_310 (null) & 0.272 & type1 & 20 \\
    512\_345 (null) & 0.150 & type1 & 5 \\
    \bottomrule
  \end{tabular}
\end{table}

\paragraph{Results.} All 8 qualifying real clusters achieve positive steering
scores (Table~\ref{tab:steering}). However, null cluster scores (0.150--0.358)
overlap substantially with those of most qualifying real clusters, limiting
the discrimination this strand provides in aggregate. The most informative
result is 768\_596 (0.419), the only cluster where a significant barycentric
predictive advantage and a positive steering score converge. Their joint occurrence in a single cluster is the
strongest functional evidence we find. For the remaining qualifying real
clusters, we do not treat steering scores as confirmatory evidence on their own.

\section{Highlight Results}
\label{sec:highlights}

\subsection{Cluster 768\_596: Convergence of Prediction and Causal Intervention}
\label{sec:highlight_596}

\begin{figure}[t]
  \centering
  \includegraphics[width=0.55\textwidth]{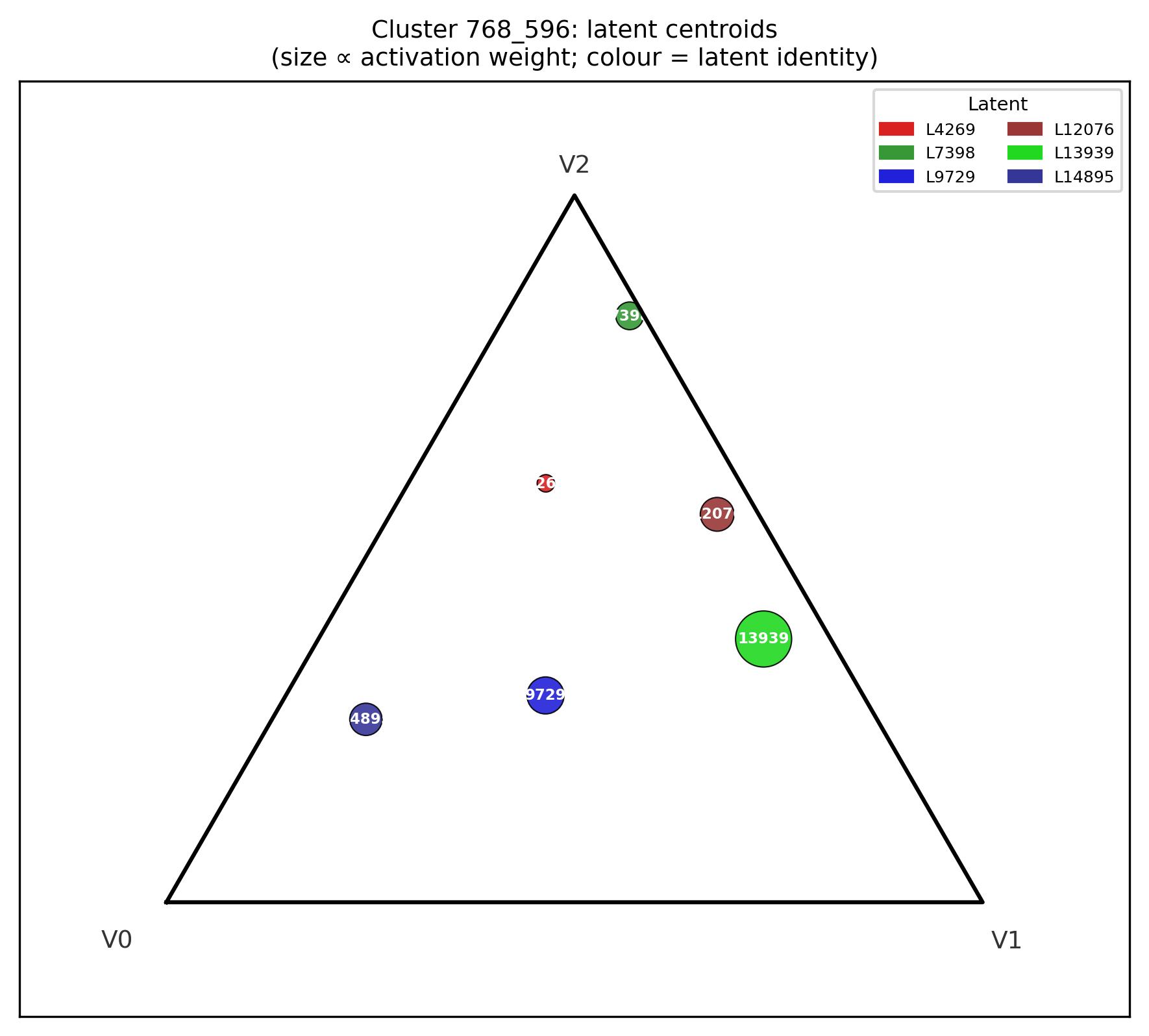}
  \caption{%
    \textbf{Cluster 768\_596: per-latent centroid positions.}
    Mean barycentric centroid of each of the six latents. The latents partition
    across the three vertices, consistent with vertex-specialized feature coding.
  }
  \label{fig:cluster_596_spatial}
\end{figure}

Cluster 768\_596 ($m = 6$ latents) has six latent centroids that partition into
three vertex-preferring groups (Figure~\ref{fig:cluster_596_spatial}). The
semantic interpretation points toward grammatical person as the organizing
dimension: V0 corresponds to third-person or external-entity reference, V1 to
first-person or self-referential framing, and V2 to second-person address and
directive language. Each vertex is primarily supported by one or two dominant
latents (Appendix~\ref{app:latent_catalogue}).

Two qualifications are important for interpreting the results. First, V2 is a
\emph{phantom vertex}: its document accuracy is 0.069, meaning the evaluator
cannot reliably identify V2-typical content even in unsteered samples. The
majority of near-vertex examples (8,258 of 10,536) land near V2, consistent
with V2 representing a high-frequency default state rather than a semantically
crisp pole. The real causal signal is in the V0$\leftrightarrow$V1 directions
(steering accuracy 0.667 and 0.429 respectively). Second, the barycentric
predictive advantage holds on near-vertex samples but does not generalize to
simplex-interior positions (interior: frac.\ bary wins $= 0.16$, mean $R^2$:
$-0.003$ vs.\ $0.093$ for the best latent), suggesting the mixture encoding is
concentrated near the vertex extremes.

With those caveats, 768\_596 is nevertheless the only cluster where passive
prediction and causal steering converge. Both signals are on the V0$\leftrightarrow$V1
axis, which represents the semantically crisp part
of the geometry. The near-vertex barycentric advantage ($p < 10^{-14}$) shows
that V0 and V1 together encode information that neither latent captures alone.
The steering score of 0.419 confirms that intervening on this axis shifts model
behavior in the expected direction.

\subsection{Cluster 512\_181: Strongest Predictive Signal Across Both Sample Regimes}
\label{sec:highlight_181}

Cluster 512\_181 ($m = 26$ latents) achieves the most unambiguous barycentric
predictive advantage in the dataset, replicating across both evaluation splits:
frac.\ bary wins $= 1.00$ on near-vertex samples ($p < 10^{-15}$, mean $R^2$:
0.612 vs.\ 0.539) and $1.00$ on simplex-interior samples ($p \approx 0$).
The replication across near-vertex and interior regimes is distinctive: whereas
768\_596's advantage is specific to the vertex extremes, 512\_181's barycentric
coordinates carry predictive signal across the full range of simplex positions.

This cluster does not qualify for causal steering evaluation (at least one
vertex has document accuracy $< 0.15$), and its semantic confidence is rated
LOW. The co-existence of a strong, replicating predictive signal with weak semantic
interpretability raises two competing interpretations. One is that the
barycentric coordinates encode a functional variable that is real but
manifests through distributional statistics rather than lexical content,
making it difficult to label from short continuations. Another is that the
26-latent cluster captures multiple overlapping contextual distinctions rather
than a single simplex mixture, producing a strong predictive signal through
aggregation rather than genuine belief-state encoding. The structured
evaluation we outline in Section~\ref{sec:future_validation} would be needed
to distinguish these.

\section{Discussion}
\label{sec:discussion}

\subsection{What the Barycentric Test Tells Us}
\label{sec:convergent}

The 5-versus-0 real-versus-null outcome is the central result. Priority clusters
were selected on geometric grounds before any functional validation, so the
contrast is not attributable to selection bias. For the five passing clusters,
the simplex geometry encodes mixture information no single feature captures,
directly ruling out the tiling artifact.

The four clusters passing interior only (512\_471, 512\_504) and the one passing
near-vertex only (768\_596) differ in \emph{where} the mixture signal lives,
not whether it exists. Clusters failing entirely, including 768\_140 (strongest
KL signal, ratio 1.987), likely encode discrete distinctions already
well-captured by a dominant individual latent. The barycentric test is the
filter that separates candidates worth deeper investigation from those that are not.

\subsection{Toward Structured Validation of Belief Geometries}
\label{sec:future_validation}

The findings establish that the pipeline finds real structure and that the
barycentric test discriminates real candidates from null. They do not, however,
confirm the belief-state interpretation that simplex coordinates encode a genuine
probability distribution over discrete latent states the model tracks as a hidden
variable.

Prior work on geometric representations relied on structured, verifiable problems
where the latent variable's value is known, allowing representations to be probed
against ground truth. Natural language discourse structure lacks these properties:
labels like ``grammatical person'' or ``referential specificity'' are not
discretely verifiable from a single token. This paper is a first attempt to extend
the search to this more abstract setting, and the gap it exposes is precisely the
absence of the structured, verifiable datasets that made prior work tractable.

Closing this gap requires datasets where simplex-position labels are assignable
reliably, enabling direct confirmation of whether the model uses belief
geometries to track underlying abstract states. Building such datasets is the
concrete next step this work motivates.

\subsection{Limitations}
\label{sec:limitations}

\paragraph{Effect sizes are modest.} The best steering score (0.419) is well
below 1.0, and steering null cluster scores overlap with most real clusters,
limiting its discrimination. Barycentric $R^2$ margins are stronger for the
clearest clusters (e.g.\ 512\_181: 0.235 vs.\ 0.160; 512\_504: 0.441 vs.\
0.307) but modest for others, consistent with simplex structures being present
but partially encoded.

\paragraph{Phantom vertices and dissociated signals.} As discussed in
Section~\ref{sec:highlight_596}, V2 of 768\_596 has document accuracy 0.069
and its near-vertex examples correspond to a default rather than a semantically
crisp state. Phantom vertices are likely a common feature of unsupervised simplex
fitting on naturalistic text and may affect other clusters not highlighted here.

\paragraph{No causal proof of belief-state tracking.} Our evidence is
correlational (barycentric $R^2$, KL) or weakly causal (steering). The
findings are consistent with belief-like state tracking but do not rule out
alternative explanations such as correlated surface features spanning the
fitted simplex.

\paragraph{Single model and layer.} All real-model results are from Gemma-2-9B,
layer 20. Whether findings generalize to other models, architectures, or layers
is unknown.

\section{Conclusion}
\label{sec:conclusion}

We have introduced and validated a pipeline for discovering simplex-structured
representational geometries in SAE latent spaces of language models. In a
controlled toy model the pipeline recovers known simplex structures. Applied
to Gemma-2-9B, the barycentric predictive advantage test discriminates real
candidates from null: 5 of 13 priority clusters pass on at least one evaluation
split; no null cluster passes on either. Cluster 768\_596 is the only case
where prediction and causal steering jointly support the simplex geometry.

These findings constitute preliminary evidence that Gemma-2-9B's SAE latent
space contains subspaces encoding genuine mixture representations of
functionally distinct latent states. The principal remaining obstacle is the
absence of structured, verifiable evaluation datasets for naturalistic discourse
variables. Future work should construct such datasets and extend the pipeline
to additional layers and model families.

\section{Reproducibility}
\label{sec:reproducibility}
The code is available at \url{https://github.com/good-epic/finding-belief-geometries}.
Submit issues on github or email the author with questions or problems.

The author leveraged LLMs in the following ways to complete this work:
\begin{itemize}
    \item \textit{Literature review and background research}
    \item \textit{Learning new concepts}. For example, I was not familiar with optimal
          transport theory before this project and used both reading and Q\&A with
          LLMs to learn about it.
    \item \textit{Programming assistance}. I wrote inline with Cursor and leveraged Claude Code
          to write code from scratch after extensive specification. I reviewed all code
          to ensure it was correct and functional.
    \item \textit{Drafting the paper}. I interactively developed an outline and first draft with
          Claude Code, then extensively edited and revised. I subsequently asked for skeptical
          rereads as well.
\end{itemize}


\bibliographystyle{colm2026_conference}
\bibliography{references}

@inproceedings{wang2009ksub,
  title={K-Subspace Clustering},
  author={Wang, Dingding and Ding, Chris and Li, Tao},
  booktitle={Machine Learning and Knowledge Discovery in Databases (ECML PKDD 2009)},
  series={Lecture Notes in Computer Science},
  volume={5782},
  pages={506--521},
  year={2009},
  publisher={Springer}
}

@article{damle2018copr,
  title={Simple, direct and efficient multi-way spectral clustering},
  author={Damle, Anil and Minden, Victor and Ying, Lexing},
  journal={Information and Inference: A Journal of the IMA},
  volume={8},
  number={1},
  pages={181--203},
  year={2019},
  publisher={Oxford University Press},
  note={Advance access published 27 June 2018},
  doi={10.1093/imaiai/iay008}
}

@misc{vanDijk2019AAnet,
  title={Finding Archetypal Spaces Using Neural Networks},
  author={van Dijk, David and Burkhardt, Daniel B. and Amodio, Matthew and Tong, Alexander and Wolf, Guy and Krishnaswamy, Smita},
  year={2019},
  eprint={1901.09078},
  archivePrefix={arXiv},
  primaryClass={cs.LG}
}

@misc{vanDijk2024AAnet,
  title={{AAnet} resolves a continuum of spatially-localized cell states to unveil tumor complexity},
  author={Venkat, Aarthi and Youlten, Scott E. and {San Juan}, Beatriz P. and Purcell, Carley and Amodio, Matthew and Burkhardt, Daniel B. and Benz, Andrew and Holst, Jeff and McCool, Cerys and Mollbrink, Annelie and Lundeberg, Joakim and van Dijk, David and Goldstein, Leonard D. and Kummerfeld, Sarah and Krishnaswamy, Smita and Chaffer, Christine L.},
  year={2024},
  howpublished={bioRxiv preprint},
  note={doi:10.1101/2024.05.11.593705}
}

@article{marzen2017predictive,
  title={Predictive rate-distortion for infinite-order {Markov} chains},
  author={Marzen, Sarah E. and Crutchfield, James P.},
  journal={Journal of Statistical Physics},
  volume={168},
  number={6},
  pages={1312--1339},
  year={2017},
  publisher={Springer}
}

@inproceedings{shai2024transformers,
  title={Transformers Represent Belief State Geometry in their Residual Stream},
  author={Shai, Adam S. and Marzen, Sarah E. and Teixeira, Lucas and
          Gietelink Oldenziel, Alexander and Riechers, Paul M.},
  booktitle={Advances in Neural Information Processing Systems},
  year={2024},
  note={arXiv:2405.15943}
}

@article{piotrowski2025constrained,
  title={Constrained Belief Updates Explain Geometric Structures in Transformer Representations},
  author={Piotrowski, Mateusz and Riechers, Paul M. and Filan, Daniel and Shai, Adam S.},
  year={2025},
  archivePrefix = {arXiv},
  eprint = {2502.01954}
}

@article{riechers2025quantum,
  title={Neural networks leverage nominally quantum and post-quantum representations},
  author={Riechers, Paul M. and Elliott, Thomas J. and Shai, Adam S.},
  year={2025},
  archivePrefix = {arXiv},
  eprint = {2507.07432}
}

@article{bricken2023monosemanticity,
  title={Towards Monosemanticity: Decomposing Language Models with Dictionary Learning},
  author={Bricken, Trenton and Templeton, Adly and Batson, Joshua and Chen, Brian
          and Jermyn, Adam and Conerly, Tom and Turner, Nick and Anil, Cem
          and Denison, Carson and Askell, Amanda and others},
  journal={Transformer Circuits Thread},
  year={2023},
  url={https://transformer-circuits.pub/2023/monosemanticity/}
}

@article{cunningham2023sparse,
  title={Sparse Autoencoders Find Highly Interpretable Features in Language Models},
  author={Cunningham, Hoagy and Ewart, Aidan and Riggs, Logan and Huben, Robert
          and Sharkey, Lee},
  year={2023},
  archivePrefix = {arXiv},
  eprint = {2309.08600}
}

@misc{saelens2024,
  title={{SAELens}: Towards Scalable Open-Source Sparse Autoencoders},
  author={Bloom, Joseph and Bricken, Trenton and others},
  year={2024},
  howpublished={GitHub repository},
  url={https://github.com/jbloomAus/SAELens}
}

@article{cutler1994archetypal,
  title={Archetypal Analysis},
  author={Cutler, Adele and Breiman, Leo},
  journal={Technometrics},
  volume={36},
  number={4},
  pages={338--347},
  year={1994},
  publisher={Taylor \& Francis}
}

@article{elhage2022toy,
  title={Toy Models of Superposition},
  author={Elhage, Nelson and Hume, Tristan and Olsson, Catherine and Schiefer, Nicholas and Henighan, Tom and Kravec, Shauna and Hatfield-Dodds, Zac and Lasenby, Robert and Drain, Dawn and Chen, Carol and Grosse, Roger and McCandlish, Sam and Kaplan, Jared and Amodei, Dario and Wattenberg, Martin and Olah, Christopher},
  year={2022},
  archivePrefix={arXiv},
  eprint={2209.10652}
}

@inproceedings{gurnee2023language,
  title={Language Models Represent Space and Time},
  author={Gurnee, Wes and Tegmark, Max},
  booktitle={International Conference on Learning Representations},
  year={2024},
  archivePrefix={arXiv},
  eprint={2310.02207}
}

@article{gurnee2025manifolds,
  title={When Models Manipulate Manifolds: The Geometry of a Counting Task},
  author={Gurnee, Wes and Ameisen, Emmanuel and Kauvar, Isaac and Tarng, Julius and Pearce, Adam and Olah, Chris and Batson, Joshua},
  year={2026},
  archivePrefix={arXiv},
  eprint={2601.04480}
}

@article{kantamneni2025trigonometry,
  title={Language Models Use Trigonometry to Do Addition},
  author={Kantamneni, Subhash and Tegmark, Max},
  year={2025},
  archivePrefix={arXiv},
  eprint={2502.00873}
}

@inproceedings{akyurek2023icl,
  title={What Learning Algorithm Is In-Context Learning? Investigations with Linear Models},
  author={Aky{\"u}rek, Ekin and Schuurmans, Dale and Andreas, Jacob and Ma, Tengyu and Zhou, Denny},
  booktitle={International Conference on Learning Representations},
  year={2023},
  archivePrefix={arXiv},
  eprint={2211.15661}
}

@inproceedings{vonoswald2023transformers,
  title={Transformers Learn In-Context by Gradient Descent},
  author={von Oswald, Johannes and Niklasson, Eyvind and Randazzo, Ettore and Sacramento, Jo{\~a}o and Mordvintsev, Alexander and Zhmoginov, Andrey and Vladymyrov, Max},
  booktitle={International Conference on Machine Learning},
  year={2023},
  archivePrefix={arXiv},
  eprint={2212.07677}
}

@inproceedings{xie2022meta-learning,
  title={An Explanation of In-Context Learning as Implicit Bayesian Inference},
  author={Xie, Sang Michael and Raghunathan, Aditi and Liang, Percy and Ma, Tengyu},
  booktitle={International Conference on Learning Representations},
  year={2022},
  archivePrefix={arXiv},
  eprint={2111.02080}
}

@article{penedo2024fineweb,
  title={{FineWeb}: Decanting the Web for the Finest Text Data at Scale},
  author={Penedo, Guilherme and Kydl{\'i}{\v{c}}ek, Hynek and allal, Loubna Ben and Lozhkov, Anton and Mitchell, Margaret and Raffel, Colin and Von Werra, Leandro and Wolf, Thomas},
  journal={arXiv preprint arXiv:2406.17557},
  year={2024}
}

\appendix

\section{Toy Model: Proof of Concept}
\label{app:toy}

\subsection*{Generator and Setup}

We construct a multipartite (MP) process by independently drawing emissions
from five component models: three Mess3 hidden Markov models
\citep{marzen2017predictive} and two Tom Quantum (Bloch Walk) generalized
HMMs \citep{riechers2025quantum}. The three Mess3 instances have parameters
$(x, a) \in \{(0.05, 0.85),\,(0.075, 0.90),\,(0.10, 0.95)\}$; the two
Tom Quantum instances have parameters $(\alpha, \beta) \in
\{(1.51, 3.07),\,(1.99, 2.51)\}$. The joint vocabulary is
$4 \times 4 \times 3 \times 3 \times 3 = 432$ tokens.

This setup is \emph{intentionally challenging}. Each token jointly encodes
contributions from all five independent sources, creating high conditional
entropy: random guessing achieves 0.2\% top-1 accuracy, and even a
well-trained transformer achieves only modest accuracy. We chose this regime
because we expected real-model belief geometries to be partially obscured by
noise and entanglement between co-occurring features. A small transformer
($d_{\mathrm{model}} = 128$, 4 heads, 3 layers, context length 16) achieves
7.1\% top-1 accuracy, confirming meaningful learning while remaining far from
saturation.

\subsection*{Recovery Results}

We fit TopK SAEs ($K \in \{3, 4, 5, 6, 7, 8, 10, 12, 14, 16, 19, 22, 25\}$,
$d_{\mathrm{SAE}} = 256$) to each layer, apply $k$-subspace clustering to the
decoder matrix, and fit AANet with target simplex dimensions $K \in \{2,3,4,5\}$,
selecting $K$ by elbow criterion. For the Mess3-associated clusters the elbow
consistently falls at $K=3$, correctly identifying the underlying 2-simplex.
Table~\ref{tab:toy_r2} shows results from the best hyperparameter setting
(TopK $K=12$, $N=6$ clusters, layer 1). All five generative components are
recovered by a unique, non-conflicting cluster with a mean $R^2$ of 0.61 on
held-out data. This is a nontrivial result: the residual stream encodes multiple
component signals along shared linear directions (cross-component entanglement),
yet the pipeline recovers all five components with no supervision. Per-component
PCA projections and representative latent activation patterns illustrating this
entanglement and the pipeline's recovery are shown in
Appendix~\ref{app:toy_pca_clusters}.

\begin{table}[h]
  \centering
  \caption{%
    \textbf{Toy model clustering recovery ($R^2$), layer 1.}
    Each cluster is assigned to the generative component whose belief state it
    best predicts (linear regression, held-out data). All five components are
    recovered with no conflicts. $R^2$ values are modest but uniformly positive,
    consistent with genuine but partial recovery under the high-entropy
    multipartite setting.
  }
  \label{tab:toy_r2}
  \small
  \begin{tabular}{lcc}
    \toprule
    Cluster & Assigned component & $R^2$ \\
    \midrule
    0 & \texttt{tom\_quantum}    & 0.551 \\
    1 & \texttt{mess3\_2}        & 0.314 \\
    2 & \texttt{tom\_quantum\_1} & 0.577 \\
    3 & \texttt{mess3\_1}        & 0.719 \\
    4 & \texttt{mess3}           & 0.893 \\
    5 & (noise)                  & ---   \\
    \midrule
    \multicolumn{2}{l}{Mean (assigned clusters)} & 0.611 \\
    \bottomrule
  \end{tabular}
\end{table}

\subsection*{Process Details}

\paragraph{Mess3 process.}  A Mess3 HMM~\citep{marzen2017predictive} has three
hidden states $\{0, 1, 2\}$ with transition probabilities parameterized by
$(x, a)$: the self-transition probability is $a$, and the remaining probability
$1-a$ is split uniformly among the other two states.  The emission from state $s$
is token $s$ with probability $x$ and a random token with probability $1-x$.

\paragraph{Tom Quantum (Bloch Walk) process.}  The Bloch Walk GHMM~\citep{riechers2025quantum}
is parameterized by $(\alpha, \beta)$ controlling the step distribution on the Bloch
sphere.  The internal state is a unit vector on $S^2$; emissions are four tokens
corresponding to projective measurements.

\paragraph{Multipartite model vocabulary.}  The MP model generates tuples from
$\{0,\ldots,3\}^2 \times \{0,1,2\}^3$, encoded as a single integer token from
$\{0,\ldots,431\}$.

\section{Toy Model: Per-Component PCA Projections and Latent Patterns}
\label{app:toy_pca_clusters}

\begin{figure}[p]
  \centering
  \begin{subfigure}[t]{0.48\textwidth}
    \centering
    \includegraphics[width=\textwidth]{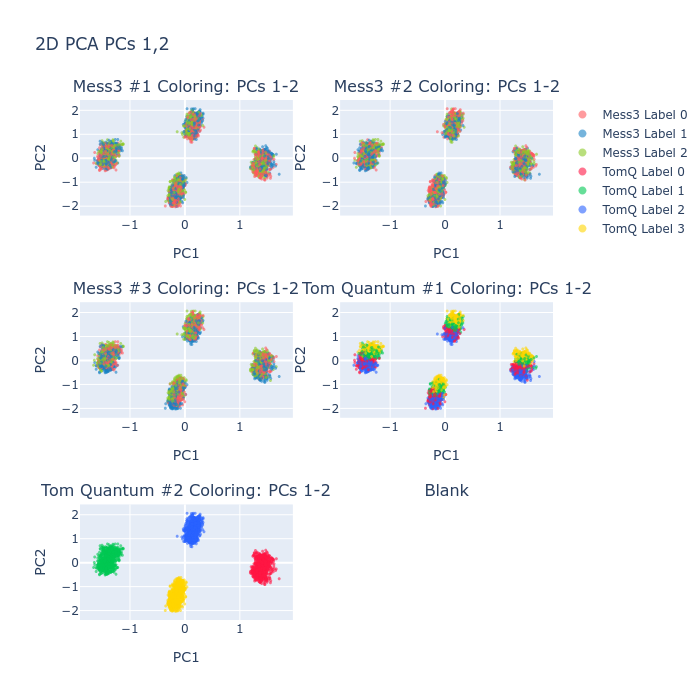}
    \caption{%
      \textbf{Tom Quantum 2, PCs 1--2.}
      TQ2's token classes are fully separated; TQ1's token classes are also
      clearly separated in this same subspace.
    }
    \label{fig:app_pca_tq2}
  \end{subfigure}
  \hfill
  \begin{subfigure}[t]{0.48\textwidth}
    \centering
    \includegraphics[width=\textwidth]{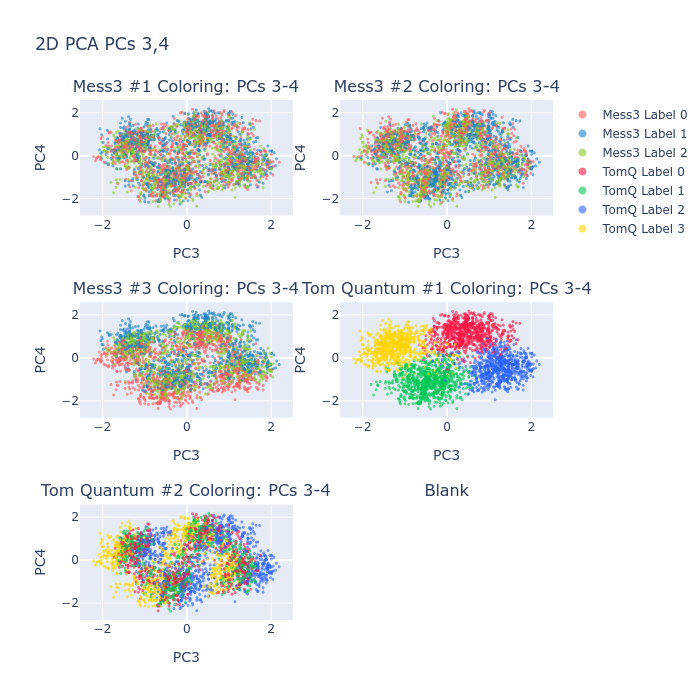}
    \caption{%
      \textbf{Tom Quantum 1, PCs 3--4.}
      TQ1's token classes separate cleanly; the same subspace also reveals
      clear separation for TQ2 and M33---three independent sub-components
      simultaneously legible in a single 2D projection.
    }
    \label{fig:app_pca_tq1}
  \end{subfigure}

  \vspace{1em}

  \begin{subfigure}[t]{0.32\textwidth}
    \centering
    \includegraphics[width=\textwidth]{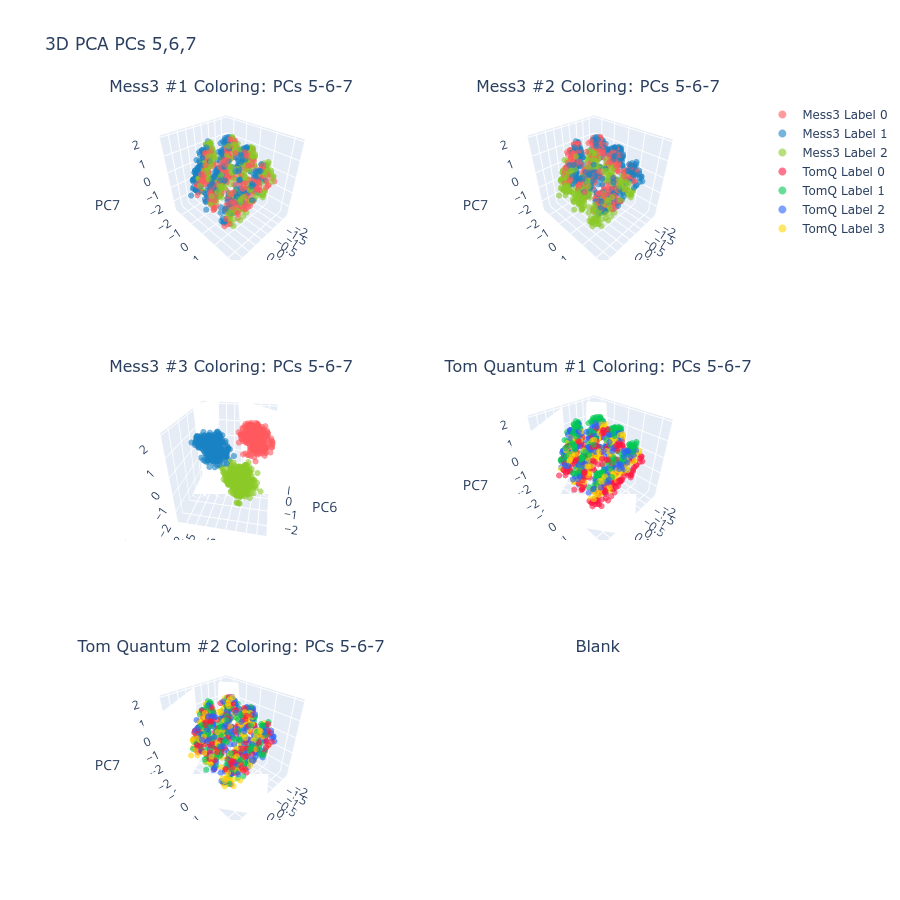}
    \caption{%
      \textbf{Mess3 \#3, PCs 5--6--7.}
      Triangular (2-simplex) structure for M33; TQ1 and M32 separation also
      visible.
    }
    \label{fig:app_pca_m33}
  \end{subfigure}
  \hfill
  \begin{subfigure}[t]{0.32\textwidth}
    \centering
    \includegraphics[width=\textwidth]{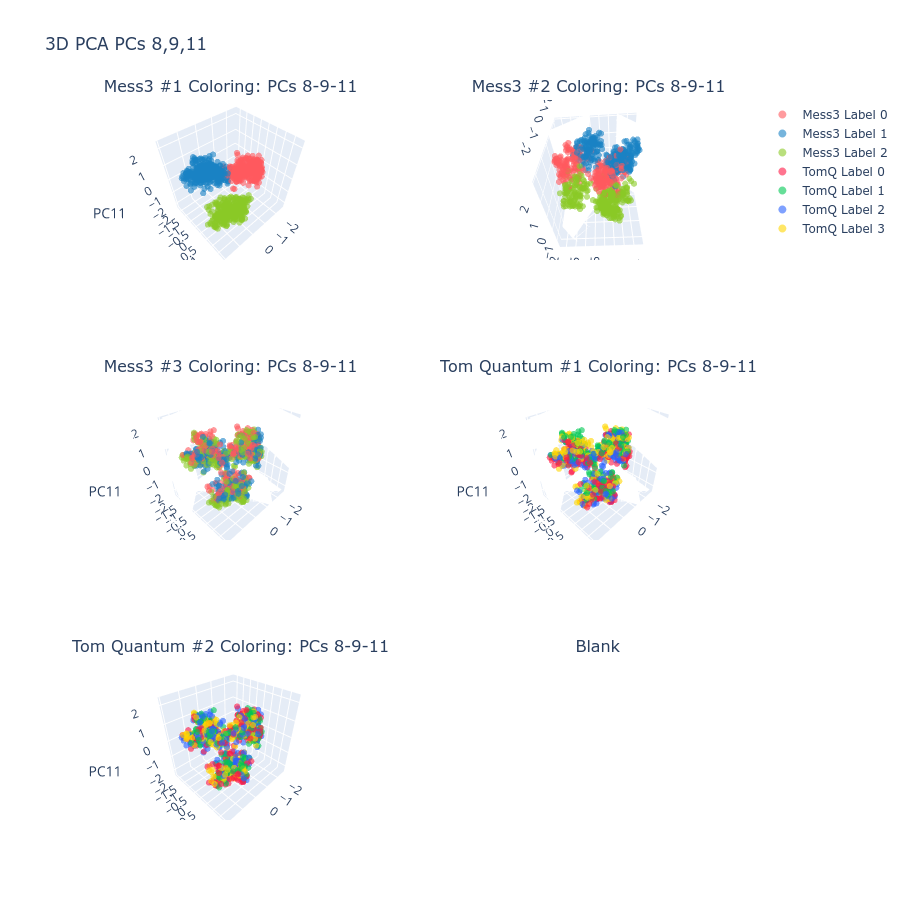}
    \caption{%
      \textbf{Mess3 \#1, PCs 8--9--11.}
      M31's triangular geometry; M32's classes are near-perfectly separated
      in this same subspace.
    }
    \label{fig:app_pca_m31}
  \end{subfigure}
  \hfill
  \begin{subfigure}[t]{0.32\textwidth}
    \centering
    \includegraphics[width=\textwidth]{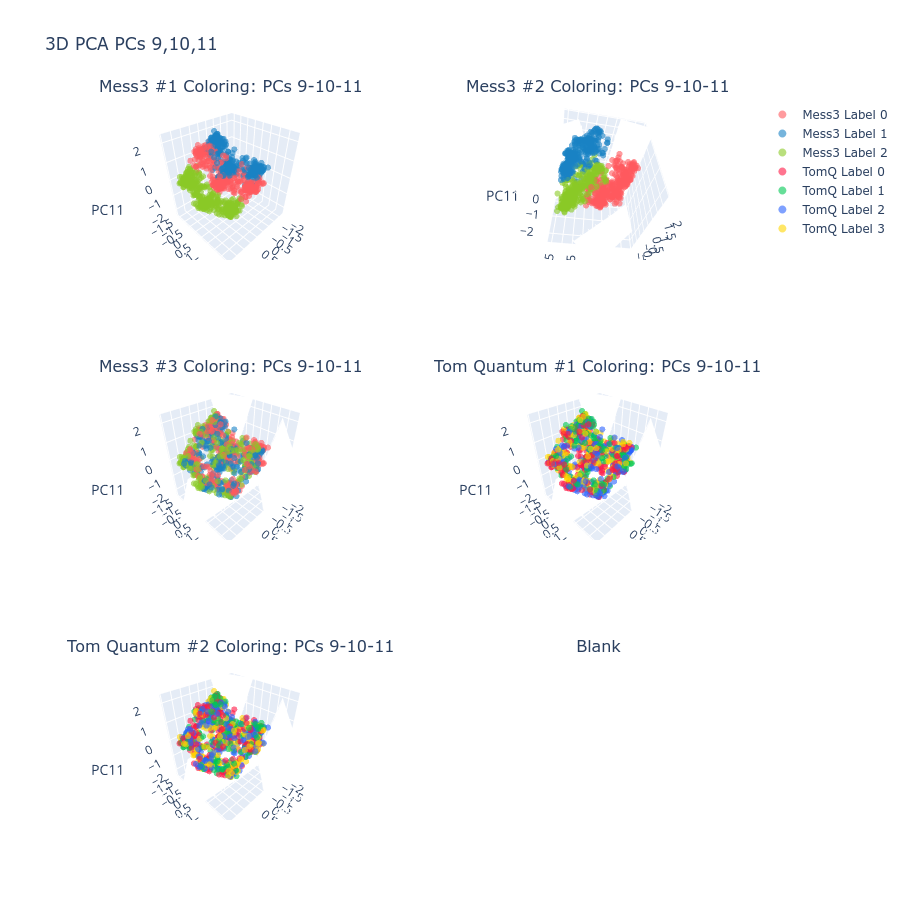}
    \caption{%
      \textbf{Mess3 \#2, PCs 9--10--11.}
      M32's triangular geometry; M31's classes are simultaneously
      near-perfectly separated.
    }
    \label{fig:app_pca_m32}
  \end{subfigure}

  \caption{%
    \textbf{Per-sub-component PCA projections, toy model layer 1.}
    Each panel shows the PCA subspace that best reveals one sub-component's
    geometry, colored by that component's true discrete output token. Every
    panel also exhibits clear separation for at least one other sub-component,
    demonstrating cross-component entanglement.
  }
  \label{fig:app_toy_cluster_pca}
\end{figure}

Figure~\ref{fig:app_toy_cluster_pca} shows the best PCA projections for each of
the five generative components at layer 1, colored by ground-truth belief state.
Different components are best revealed by different principal components. More
strikingly, every panel also exhibits clear token-class separation for at least
one other sub-component, demonstrating cross-component entanglement: the residual
stream encodes multiple independent generative sources along the same linear
directions. The TQ1 projection (PCs 3--4) simultaneously reveals clear separation
for TQ2 and M33; the M31 and M32 projections each show near-perfect separation
for the other. Despite this entanglement, the recovered clusters achieve a mean
$R^2$ of 0.61 (Table~\ref{tab:toy_r2}).

\begin{figure}[t]
  \centering
  \includegraphics[width=\columnwidth]{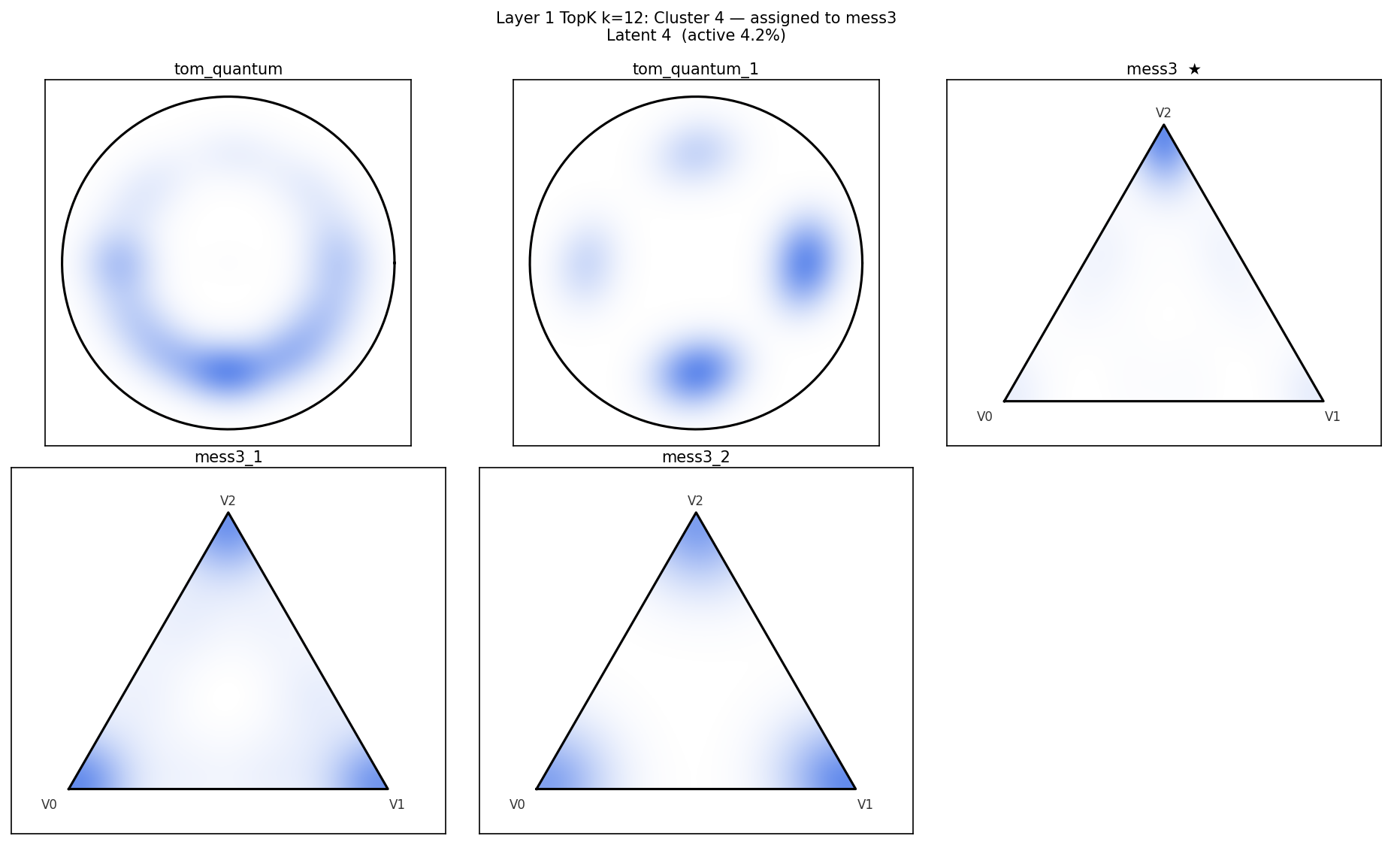}
  \vspace{0.5em}
  \includegraphics[width=\columnwidth]{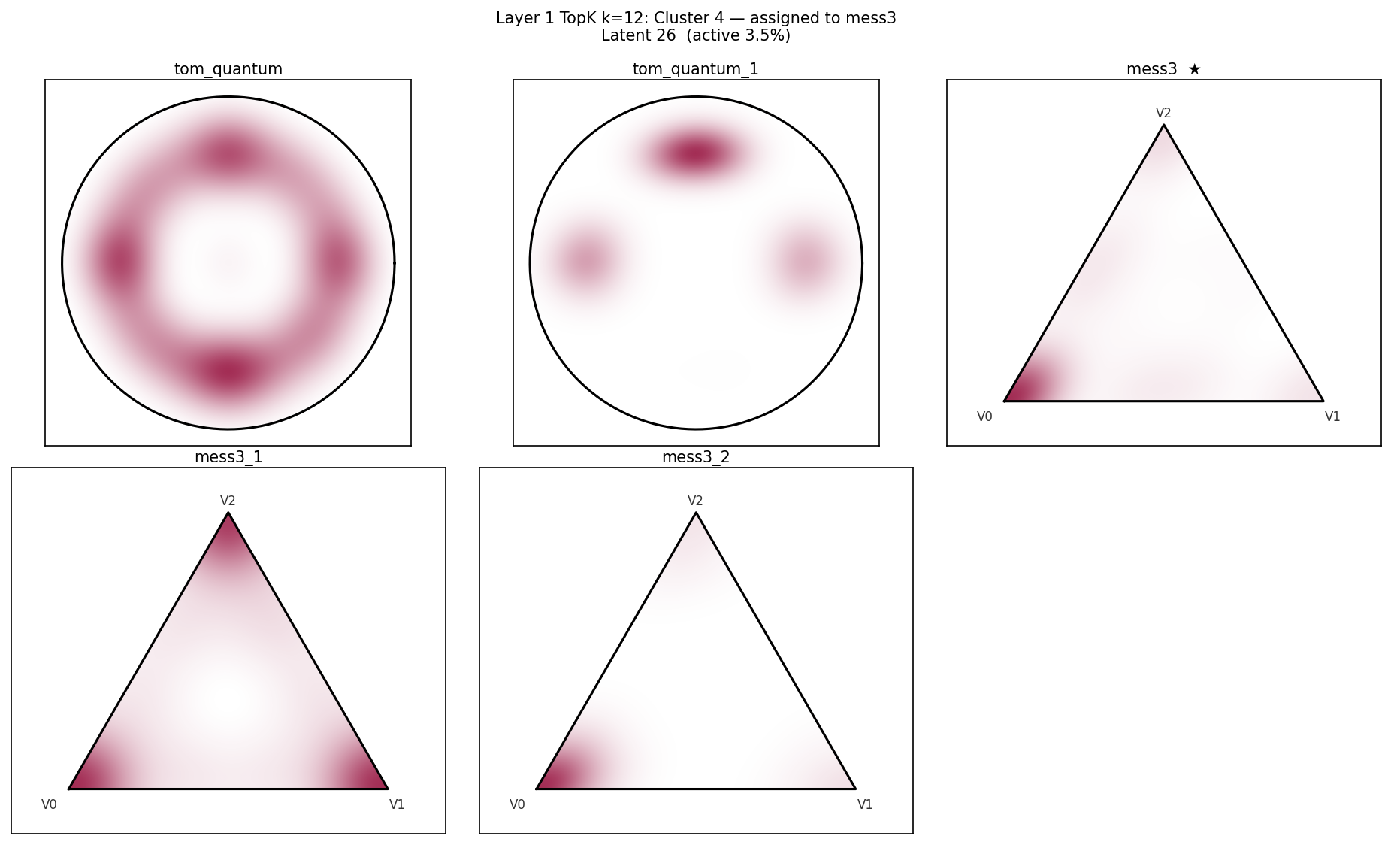}
  \caption{%
    \textbf{Two representative latents from Cluster~4 (multipartite toy model,
    layer~1, TopK $K=12$).}
    Each panel shows KDE-smoothed activation density over all five component
    belief geometries. The assigned component is marked~$\star$ ($R^2 = 0.89$).
    \emph{Top:} Latent~4 fires near the top vertex of each Mess3 simplex and
    at the centre of the Tom Quantum disks.
    \emph{Bottom:} Latent~26 fires near the base vertices and along the annular
    rim of the Tom Quantum disks.
    The two latents tile complementary regions of the same geometry,
    demonstrating geometry-consistent partitioning of belief-state information.
  }
  \label{fig:toy_epdf}
\end{figure}

Figure~\ref{fig:toy_epdf} shows the empirical probability density functions for
all latents in the best natural cluster at layer~1 (Cluster~4, $R^2 = 0.89$).
Individual latents concentrate in distinct regions of each component's belief
geometry, demonstrating geometry-consistent partitioning of belief-state
information across latents.

\section{SAE Details}
\label{app:sae}

We use the Gemma-2-9B SAE from the SAELens library (release
\texttt{gemma-scope-9b-pt-res}, layer 20, expansion factor $\approx 4.6$,
average L0 $\approx 68$ active features per token).  The SAE was trained by
DeepMind on a proprietary mixture of text data using JumpReLU activation with
$L_2$ reconstruction loss.

For the toy model SAEs, we use a custom TopK SAE implementation with
$d_{\mathrm{SAE}} = 256$, training for 20{,}000 steps on sequences generated
online from the MP model. We tried $K \in \{3, 4, 5, 6, 7, 8, 10, 12, 14, 16,
19, 22, 25\}$; the best-recovering SAE for the Mess3 component uses $K = 3$.

\section{AANet Architecture and Training}
\label{app:aanet}

We use the AANet implementation from \citet{vanDijk2019AAnet}.  The encoder
consists of two hidden layers (widths 256, 128 by default), a bottleneck of
dimension $K-1$ (to enforce the simplex constraint via softmax normalization),
and an analogous decoder.  The loss combines $L_2$ reconstruction, a simplex
penalty $\lambda_1 (1 - \|\mathbf{E}(a)\|_1)$, and a non-negativity penalty
$\lambda_2 \sum_i |x_i| \mathbf{1}(x_i < 0)$.  We train with AdamW,
learning rate $10^{-3}$, for 10{,}000 steps, selecting the checkpoint with the
lowest validation reconstruction loss, averaged across 5 independent restarts.

\section{Causal Steering Intervention Types}
\label{app:steering_types}

The three intervention types differ in how long the steering delta is sustained
during generation. In all cases the delta is the AANet-derived direction toward
the target vertex, added to the residual stream at scale $s$.

\begin{itemize}
  \item \textbf{Type 1}: Single-position patch.  The delta is added only to
    the cached residual at the trigger position; subsequent generated tokens
    attend to this modified position via the KV cache but are not directly
    patched.
  \item \textbf{Type 2}: Sustained window patch.  The delta is injected into
    each newly computed position for the first $k_{\mathrm{sustain}}$ generated
    tokens after the trigger, then generation proceeds freely.
  \item \textbf{Type 3}: Fully sustained patch.  The delta is applied to every
    generated position throughout the full continuation.
\end{itemize}

\section{Barycentric vs.\ Best-Latent $R^2$: Detailed Plots}
\label{app:bary_plots}

\begin{figure}[H]
  \centering
  \includegraphics[width=\textwidth]{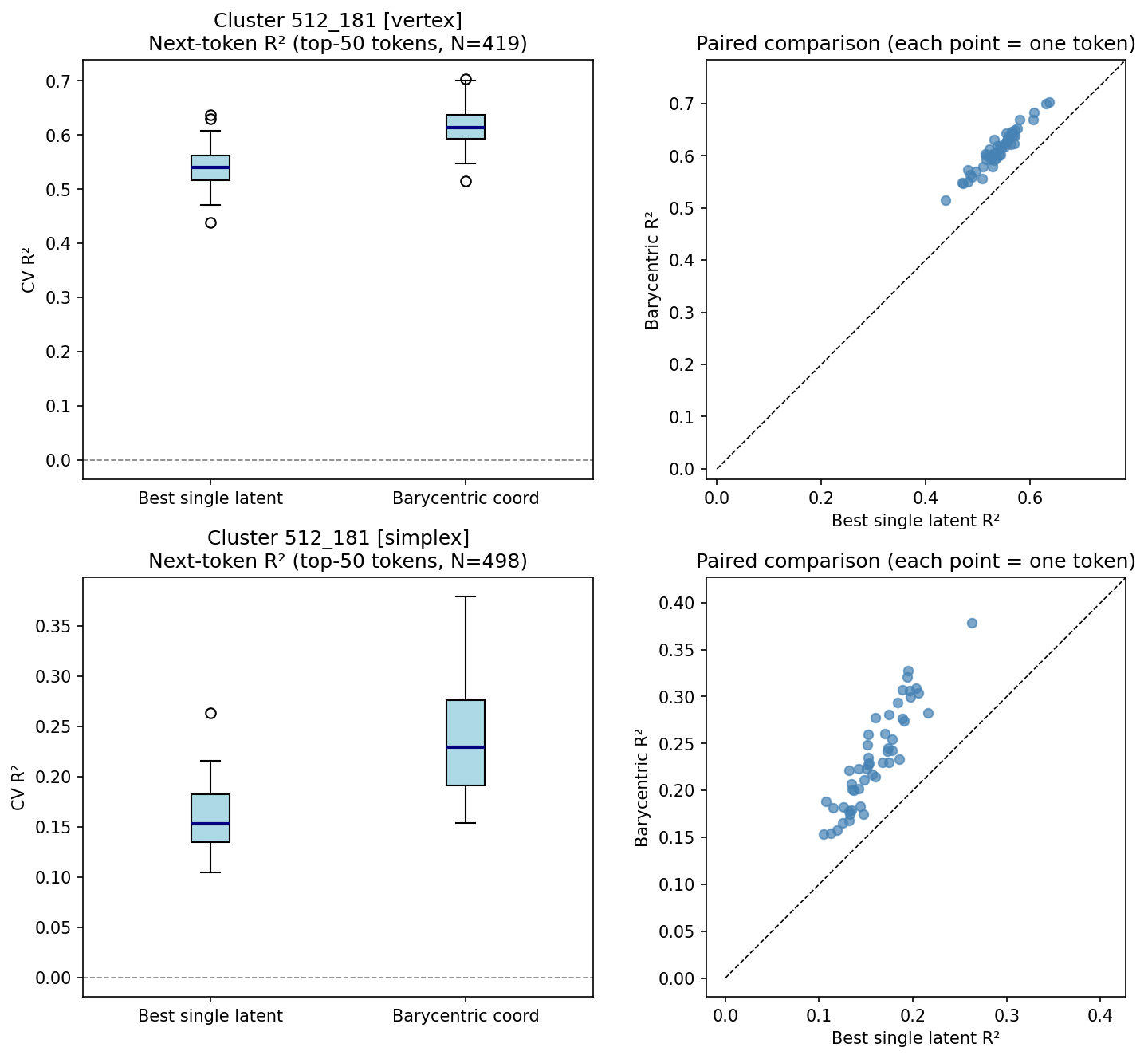}
  \caption{%
    \textbf{Cluster 512\_181: barycentric vs.\ best-latent $R^2$.}
    Barycentric coordinates (mean $R^2 = 0.612$) outperform the best individual
    latent (mean $R^2 = 0.539$) for every one of the 50 tokens
    (Wilcoxon $p < 10^{-15}$).
  }
  \label{fig:app_bary_181}
\end{figure}

\begin{figure}[H]
  \centering
  \includegraphics[width=\textwidth]{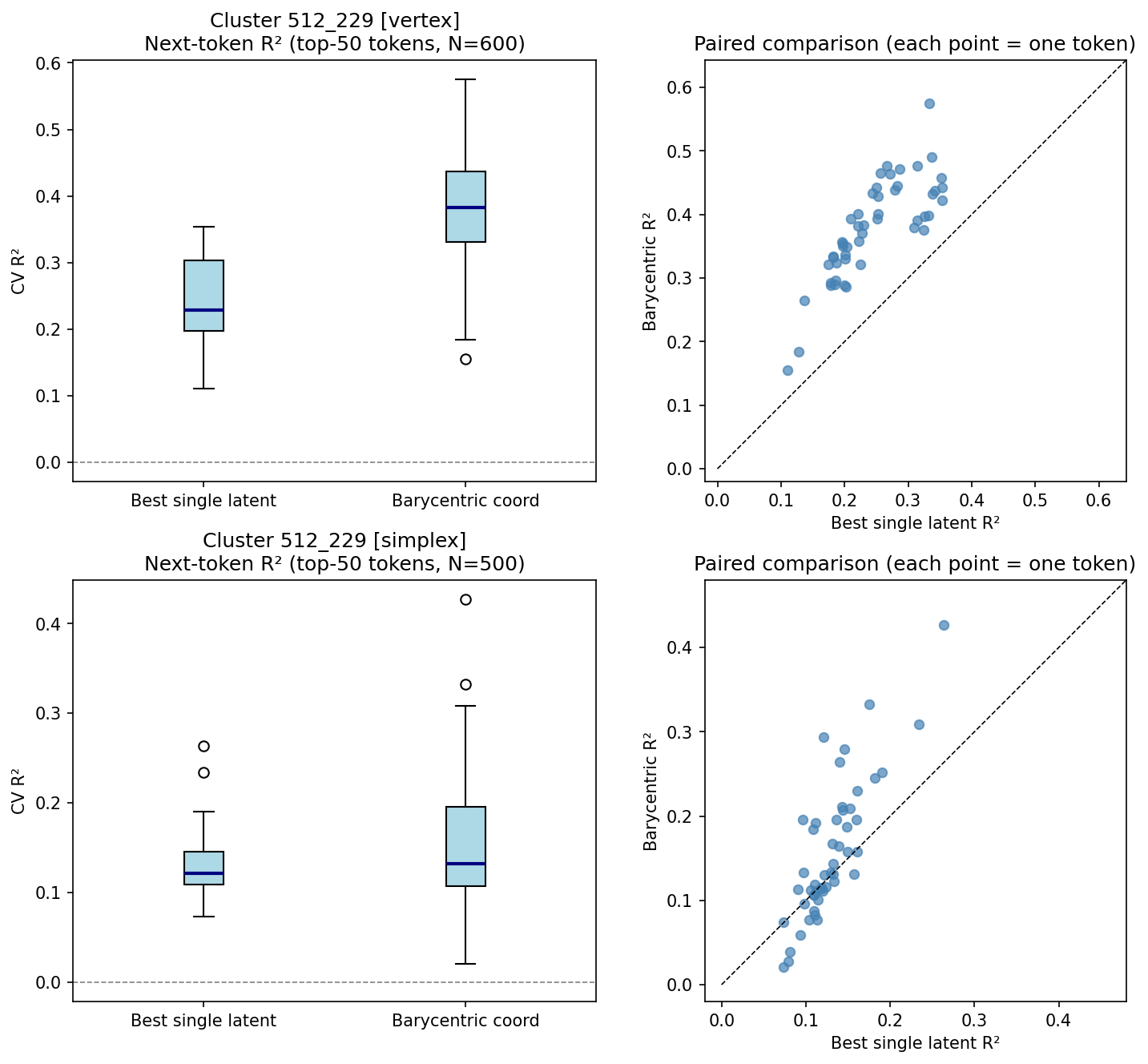}
  \caption{%
    \textbf{Cluster 512\_229: barycentric vs.\ best-latent $R^2$.}
    Barycentric coordinates (mean $R^2 = 0.378$) outperform the best individual
    latent (mean $R^2 = 0.244$) for all 50 tokens (Wilcoxon $p < 10^{-15}$),
    with a $\sim$55\% relative improvement.
  }
  \label{fig:app_bary_229}
\end{figure}

\begin{figure}[H]
  \centering
  \includegraphics[width=\textwidth]{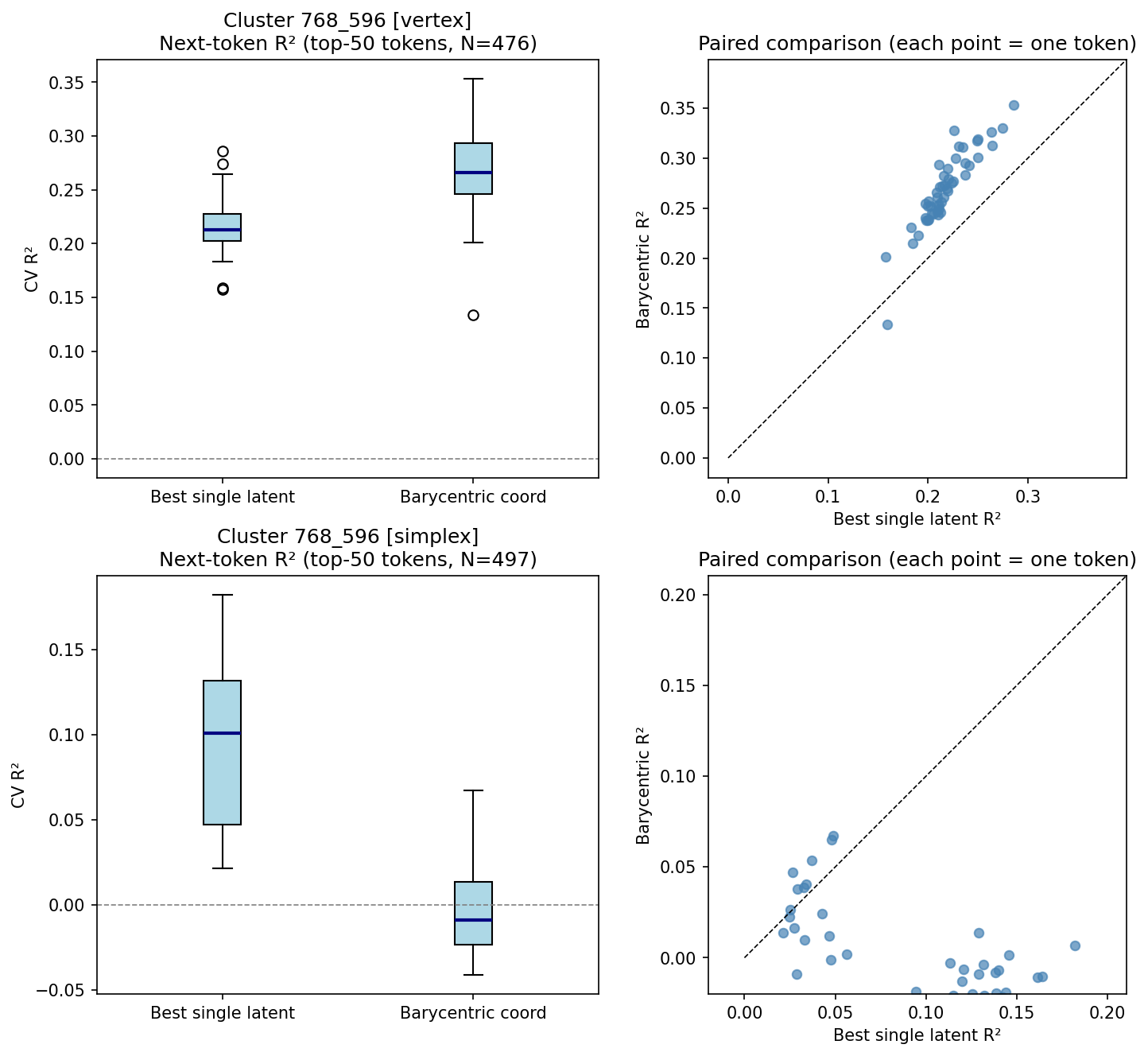}
  \caption{%
    \textbf{Cluster 768\_596: barycentric vs.\ best-latent $R^2$.}
    Barycentric coordinates (mean $R^2 = 0.269$) outperform the best individual
    latent (mean $R^2 = 0.217$) for 98\% of tokens (Wilcoxon $p < 10^{-14}$).
    This cluster also has the highest causal steering score (0.419).
  }
  \label{fig:app_bary_596}
\end{figure}

Figures~\ref{fig:app_bary_181}--\ref{fig:app_bary_596} show the full $R^2$
comparison plots for the three clusters with a significant barycentric advantage
on near-vertex samples.  Each figure contains a box plot comparing per-token
$R^2$ distributions (left) and a paired scatter plot where points above the
diagonal indicate tokens where bary wins (right).

\section{All Priority Cluster Results}
\label{app:all_results}

\begin{table}[h]
  \centering
  \caption{%
    \textbf{Full results across all four validation strands.}
    KL ratio: cross/same symmetric KL divergence ratio.
    Bary wins: fraction of top-50 tokens where $R^2_{\mathrm{bary}} >
    R^2_{\mathrm{best\,latent}}$ (``sig.'' if Wilcoxon $p < 0.001$).
    Steering: best-combination score (``---'' if cluster did not qualify).
    Interp.: frontier-LLM vertex label consistency (HIGH/MED/LOW).
  }
  \label{tab:full_results}
  \begin{tabular}{lccccc}
    \toprule
    Cluster & $m$ & KL ratio & Bary wins & Steering & Interp. \\
    \midrule
    512\_17   & 15 & 1.054 & 0.00       & 0.118 & HIGH \\
    512\_22   & 6  & 1.058 & 0.00       & 0.356 & LOW \\
    512\_67   & 25 & 1.229 & 0.00       & 0.250 & LOW \\
    512\_181  & 26 & 1.037 & 1.00 (sig.)& ---   & LOW \\
    512\_229  & 22 & 1.166 & 1.00 (sig.)& ---   & MED \\
    512\_261  & 18 & 1.032 & 0.00       & 0.242 & MED \\
    512\_471  & 10 & 1.017 & 0.00$^\dagger$  & ---   & LOW \\
    512\_504  & 22 & 1.164 & 0.00$^\dagger$  & ---   & MED \\
    768\_140  & 26 & \textbf{1.987} & 0.20 & 0.289 & MED \\
    768\_210  & 4  & 1.017 & 0.00       & 0.336 & MED \\
    768\_306  & 5  & 1.102 & 0.16       & ---   & MED \\
    768\_581  & 11 & 1.144 & 0.00       & 0.209 & LOW \\
    768\_596  & 6  & 1.037 & 0.98 (sig.)& \textbf{0.419} & MED \\
    \midrule
    512\_138 (null) & 51 & 1.329 & 0.00 & 0.358 & --- \\
    512\_345 (null) & 47 & 1.311 & 0.02 & 0.150 & --- \\
    768\_310 (null) & 17 & 1.108 & 0.16 & 0.272 & --- \\
    \bottomrule
  \end{tabular}
  \smallskip\\
  \small $^\dagger$ Bary wins 0\% on near-vertex samples but significant on
  simplex-interior samples (512\_471: 84\%, 512\_504: 100\%, both $p \approx 0$).
\end{table}

\section{Per-Cluster Latent Catalogue}
\label{app:latent_catalogue}

For each of the four clusters with the strongest functional evidence---the
highlight cluster (768\_596), the two clusters with the largest barycentric
margins (512\_181, 512\_229), and the cluster with the strongest KL signal
and semantic interpretation (768\_140)---we list every cluster latent
alongside its Neuronpedia GPT-4o-mini auto-interpretation and the simplex
vertex(es) at which it is active in near-vertex samples.

A latent is counted as \emph{active at vertex~$v$} if its mean activation
across near-vertex-$v$ samples accounts for at least 1\% of the total mean
activation summed across all cluster latents at that vertex. The \textbf{pie\%}
column reports this fractional share; a dash (---) indicates the latent did
not reach the 1\% threshold at any vertex.

\subsection*{Cluster 512\_181 \quad $k=3$, synthesis confidence: \textbf{low}}

\smallskip
\noindent\textbf{V0} (mixed): Content words and lexical items carrying semantic meaning.\\
\textbf{V1} (mixed): Function words and determiners serving grammatical roles.\\
\textbf{V2} (inconsistent): Technical/specialized terminology OR prepositions and connectives.

\smallskip
\begin{longtable}{p{1.0cm} p{7.5cm} p{1.6cm}}
\toprule
\textbf{Latent} & \textbf{Neuronpedia interp (GPT-4o-mini)} & \textbf{Active at (pie\%)} \\
\midrule
\endhead
\bottomrule
\endfoot
L2149    & Occurrences of research methodologies and results in scientific texts              & V1 (19\%) \\
L2318    & Instances of error logging and debugging messages in code                         & --- \\
L2759    & Programming constructs and function calls associated with database management     & --- \\
L3510    & Mathematical expressions related to derivatives                                   & --- \\
L3954    & Mathematical concepts related to derivatives and calculus                         & --- \\
L4321    & References to privilege and inequality in social contexts                          & V1 (73\%) \\
L4389    & References to legal principles and judicial interpretations                        & --- \\
L4741    & References to educational contexts and programming structures                      & --- \\
L6538    & Genetic and phenotypic differences between mutant and wild-type organisms          & --- \\
L7419    & Terms related to specialized or technical concepts in various fields of study      & --- \\
L7988    & Structured data, possibly focusing on code or programming attributes               & --- \\
L8095    & API request parameters and their associated values                                & --- \\
L8509    & Technical terms and references related to programming and system processes         & --- \\
L9055    & Elements related to variable names and their associated values within programming code & --- \\
L10207   & Structured elements and data types in programming code                            & V2 (49\%) \\
L11115   & Data values and statistical metrics related to medical research                    & --- \\
L11611   & Metrics and effects related to dietary treatments in clinical or experimental contexts & V1 (2\%) \\
L12475   & Structured programming syntax elements and errors                                  & V2 (3\%) \\
L14402   & Programming or coding-related syntax and structures                                & --- \\
L15310   & Elements related to mathematical operations or functions in coding                  & --- \\
L15334   & Elements related to data structures and their connections                          & V0 (97\%), V2 (48\%) \\
L15367   & References to specific legal and regulatory terms                                  & V1 (3\%) \\
L15676   & Components and notations related to mathematical and physical formulas             & --- \\
L15798   & Technical terms and structures related to electrical engineering and data processing & --- \\
L16183   & Specific mentions of maps and locations                                            & --- \\
L16184   & Mentions of numerical values or mathematical parameters                            & --- \\
\end{longtable}

\subsection*{Cluster 768\_140 \quad $k=3$, synthesis confidence: \textbf{medium}}

\smallskip
\noindent\textbf{V0} (mixed): Generic/indefinite reference (common nouns, articles, abstract concepts).\\
\textbf{V1} (consistent): Specific/definite reference (proper nouns, named entities, dates, locations).\\
\textbf{V2} (mixed): Contextual/anaphoric reference (pronouns, demonstratives, function words).

\smallskip
\begin{longtable}{p{1.0cm} p{7.5cm} p{1.6cm}}
\toprule
\textbf{Latent} & \textbf{Neuronpedia interp (GPT-4o-mini)} & \textbf{Active at (pie\%)} \\
\midrule
\endhead
\bottomrule
\endfoot
L449     & Special characters, particularly underscores and certain punctuation marks         & --- \\
L1234    & Various forms of punctuation and mathematical symbols                              & --- \\
L1367    & References to academic papers and their citation formats                           & --- \\
L1462    & Quotes or dialogue from characters or people                                       & V0 (5\%) \\
L1575    & References to experimental findings and numerical data                             & --- \\
L1993    & Punctuations and metadata references within the text                               & --- \\
L2493    & Quotes, citations, and references from authoritative figures or texts              & --- \\
L2756    & Punctuation and formatting signals in the text                                     & V0 (3\%) \\
L2878    & Emotional expressions and sentiments related to personal connections                & V1 (2\%) \\
L3771    & Direct quotations and dialogue in the text                                         & V0 (7\%) \\
L3988    & Instances of dialogue and quotation marks                                          & --- \\
L4449    & Quotations and speech indicators                                                   & V0 (3\%) \\
L4927    & Phrases that signal emphasis or specificity in statements                           & --- \\
L5030    & Elements related to JavaScript code and function definitions                        & --- \\
L6432    & Elements related to storytelling and narrative structure                            & --- \\
L7176    & Quoted speech and dialogue within the text                                         & V0 (26\%) \\
L8015    & Terms related to online services and technology                                     & V2 (1\%) \\
L8868    & Critical events and significant moments in various contexts                         & V1 (13\%) \\
L9395    & Dialogue and quotes from individuals within the text                               & V0 (6\%) \\
L10473   & Scientific terms related to medical and biological research contexts                & --- \\
L10536   & References to specific warranty policies and legal terms                            & V0 (40\%) \\
L11869   & Quotes and references to statements made by individuals in a critical context       & V0 (1\%), V1 (2\%) \\
L12387   & Phrases that indicate personal reflection or apology                               & V0 (2\%) \\
L13511   & Keywords related to medical research and development                               & V2 (97\%) \\
L15111   & Instances of emphasis or quotes around significant concepts                        & V0 (8\%) \\
L16340   & Informational or procedural text related to finance and taxes                       & V1 (82\%) \\
\end{longtable}

\subsection*{Cluster 768\_596 \quad $k=3$, synthesis confidence: \textbf{medium}}

\smallskip
\noindent\textbf{V0} (mixed): Third-person references and impersonal/descriptive constructions.\\
\textbf{V1} (mixed): First-person perspective and subjective expression.\\
\textbf{V2} (mixed): Second-person address and directive/instructional language.

\smallskip
\begin{longtable}{p{1.0cm} p{7.5cm} p{1.6cm}}
\toprule
\textbf{Latent} & \textbf{Neuronpedia interp (GPT-4o-mini)} & \textbf{Active at (pie\%)} \\
\midrule
\endhead
\bottomrule
\endfoot
L4269    & Code-related syntax and structures                                                  & --- \\
L7398    & Conversational expressions and phrases                                             & V1 (2\%), V2 (100\%) \\
L9729    & Punctuation marks and their contexts within sentences                               & --- \\
L12076   & Questions related to religious practices and their origins                          & V1 (2\%) \\
L13939   & References to non-profit organizations and charitable donations                      & V1 (97\%) \\
L14895   & Code-related terms and programming concepts                                         & V0 (99\%) \\
\end{longtable}

\subsection*{Cluster 512\_229 \quad $k=3$, synthesis confidence: \textbf{medium}}

\smallskip
\noindent\textbf{V0} (consistent): Simple syntactic structures: isolated words, basic phrases, common function words.\\
\textbf{V1} (mixed): Formal/technical content: proper nouns, dates, specialized terminology.\\
\textbf{V2} (consistent): Complex syntactic structures: dense modification, embedded clauses.

\smallskip
\begin{longtable}{p{1.0cm} p{7.5cm} p{1.6cm}}
\toprule
\textbf{Latent} & \textbf{Neuronpedia interp (GPT-4o-mini)} & \textbf{Active at (pie\%)} \\
\midrule
\endhead
\bottomrule
\endfoot
L1189    & Snippets of mathematical or programming expressions and logical statements          & --- \\
L1277    & References to individuals in positions of authority or influence                    & --- \\
L1425    & References to historical events and their impacts on society                        & --- \\
L1481    & References to educational institutions and their operational details                 & V1 (16\%) \\
L1540    & Words indicative of financial or economic conditions and events                      & V1 (2\%) \\
L1818    & Punctuation and markup elements in text                                             & --- \\
L3373    & Specific dates and numbers, particularly related to events or data points           & V1 (15\%) \\
L5990    & References to specific scientific publications and data repositories                 & --- \\
L6370    & Specific numerical data and dates associated with events                             & V1 (7\%) \\
L7478    & References to specific locations, organizational names, and demographic details      & V1 (1\%) \\
L7833    & Numerical and procedural references in formal documents and academic contexts        & V0 (100\%), V2 (100\%) \\
L8269    & Occurrences of specific dates and times                                             & V1 (11\%) \\
L8501    & Keywords related to financial burdens and energy-related issues                      & --- \\
L8625    & Time-related terms and dates                                                        & V1 (5\%) \\
L9122    & Statements related to programming constructs and their syntax                        & V1 (1\%) \\
L9805    & Economic indicators related to inflation and consumer behavior                       & V1 (2\%) \\
L13248   & Patterns in structured data or code, relating to token identifiers or variables      & --- \\
L13972   & Details related to historical events and patent information                         & --- \\
L14155   & Numbers and financial figures related to statistics                                  & V1 (31\%) \\
L14760   & Specific dates and numerical data                                                   & V1 (2\%) \\
L15855   & Numerical values and specific identifiers related to data points                    & V1 (5\%) \\
L16151   & Indicators of conditional statements and significance                               & --- \\
\end{longtable}

\end{document}